\pdfoutput=1

\documentclass[11pt]{article}

\usepackage{acl}

\usepackage{times}
\usepackage{latexsym}
\usepackage{tcolorbox}
\usepackage{enumitem} %
\usepackage{tikz}
\usepackage{tabularx}
\usetikzlibrary{positioning,calc} 

\usepackage[T1]{fontenc}

\usepackage[utf8]{inputenc}

\usepackage{microtype}

\usepackage{inconsolata}

\usepackage{graphicx}
\usepackage{booktabs} %
\usepackage{subcaption}

\usepackage{xcolor}

\title{The Rarity Blind Spot: A Framework for Evaluating Statistical Reasoning in LLMs}

\author{Seiji Maekawa \\
Megagon Labs\\
\texttt{seiji@megagon.ai} \And
Hayate Iso\thanks{~~Work done while at Megagon Labs.} \\
NVIDIA\\
\texttt{hiso@nvidia.com} \And
Nikita Bhutani \\
Megagon Labs\\
\texttt{nikita@megagon.ai}
}

\begin{document}
\maketitle
\begin{abstract}

Effective decision-making often relies on identifying what makes each candidate distinctive. While existing benchmarks for LLMs emphasize retrieving or summarizing information relevant to a given query, they do not evaluate a model's ability to identify globally distinctive features across a set of documents. We introduce Distinctive Feature Mining (DFM), 
a new task that challenges models to analyze a small-to-medium collection (10-40 documents) and surface features that are rare in the global context (e.g., appearing in less than 10\% of documents). This setting mirrors real-world scenarios such as candidate selection or product differentiation, where statistical reasoning, not retrieval, is key. 
To enable systematic evaluation of this capability, we present \textsc{DiFBench}, a configurable benchmark creation framework with controllable parameters such as document set size and distinctiveness thresholds.

Using \textsc{DiFBench}, we perform a large-scale assessment of distinctive feature mining across ten state-of-the-art LLMs. Our findings reveal a significant performance gap between general-purpose and reasoning-enhanced models. All models, however, substantially degrade as the task complexity and document count increase. 
We also find that a common failure mode is misidentifying frequent features as distinctive. 
These insights reveal core limitations in contemporary LLMs' abilities to perform fine-grained, statistical reasoning and rarity detection. Our code and data are available: \url{https://github.com/megagonlabs/DiFBench}

\end{abstract}

\begin{figure}
\centering
\includegraphics[width=.94\linewidth]{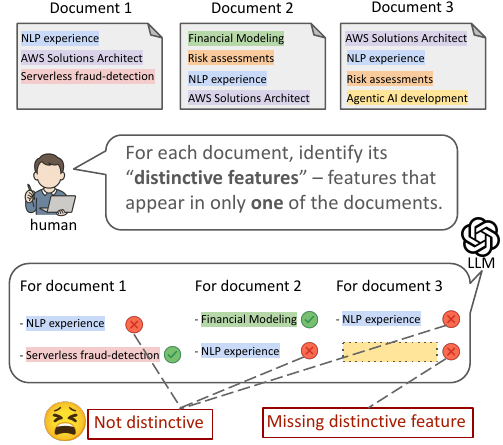}
\caption{Example of Distinctive Feature Mining (DFM). Given a set of documents, the model needs to identify globally rare features. Here, the model incorrectly identifies \textit{``NLP experience''} as distinctive, when it is shared by all documents. In contrast, it misses the truly rare feature \textit{``Agentic AI development''}.}
\label{fig:dfm_example}
\end{figure}

\section{Introduction}
\label{sec:intro}

When making decisions from large candidate pools--whether selecting products, evaluating applicants, or analyzing documents--humans naturally seek to understand what makes each candidate distinctive.  This cognitive process of identifying uncommon or unique traits is central to effective decision-making. As LLMs are increasingly deployed in recommendation and decision support systems across domains such as hiring \cite{an-etal-2024-large,iso-etal-2025-evaluating} and travel planning \cite{xie2024travelplanner}, their ability to mimic this core human capability becomes critical.

Our investigation reveals a fundamental limitation: even state-of-the-art reasoning models fail to recognize rarity when analyzing a set of documents. For instance, when analyzing technical resumes, a model might mistakenly identify \textit{``NLP experience''} as distinctive when it is shared by multiple documents, and yet miss genuinely rare skills like \textit{``Agentic AI development''} (see Figure \ref{fig:dfm_example}). This behavior is akin to the psychological phenomenon of 
\textit{base rate neglect}~\cite{tversky1974judgment,grether2012seeing}, where statistical frequency is ignored in favor of more salient but less informative cues. This can lead to systematically suboptimal recommendations and decisions.

LLM benchmarks have primarily focused on query-driven tasks, such as sparse information retrieval (e.g., the needle-in-a-haystack test \cite{kamradt2024niah}) or multi-document and long-context reasoning \cite{karpinska-etal-2024-one,xu2024stress,kuratov2024babilong,levy-etal-2024-task,bai-etal-2024-longbench,zhang-etal-2024-bench,hsieh2024ruler,bai2024longbench2,yen2025helmet,maekawa2025holistic}. 
These benchmarks assess a model’s ability to find or aggregate relevant information, often in response to an explicit query. However, they do not test whether a model can derive global statistical insights across a collection, in particular those involving feature rarity.

To fill this gap, we introduce \textbf{Distinctive Feature Mining (DFM)}, a new task that requires identifying globally rare attributes (appearing in $\leq\theta$\% of documents) within document collections. Unlike traditional retrieval or summarization, DFM requires \textit{statistical reasoning} over a population, not just extracting salient information from individual documents. We focus on collections of 10–40 documents, a realistic scale for decisions like candidate screening or product comparison. This scale is large enough to require aggregate reasoning and base-rate estimation, yet small enough to demand holistic comprehension and accurate attribution.

We operationalize this through \textbf{DiFBench}, a configurable benchmark creation framework that precisely governs feature distributions. For example, it ensures \textit{``blockchain development''} appears in 2 out of 40 resumes (5\%) while \textit{``project management''} appears in 25 (62.5\%). This  enables systematic evaluation across document scales and domains, with controllable parameters including document count (10-40), feature density, and distinctiveness thresholds (2.5\%-20\%).

Our evaluation over 10 state-of-the-art LLMs reveals three key findings: (1) non-reasoning models achieve F1 < 30\%, revealing limitations in multi-document reasoning; (2) even advanced models (o3, Gemini-2.5-Flash) degrade from F1 > 85\% on 10 documents to F1 < 60\% on 40 documents; and (3) 75.9\% of errors involve misclassifying common features as distinctive. This precision drop mirrors base rate neglect in human cognition. We mitigate this via explicit verification prompting, achieving a 65\% relative F1 gain while maintaining recall.

The main contributions of this work include:
\begin{enumerate}[label=(\arabic*), leftmargin=*]
    \item We introduce \textbf{DFM} task and \textbf{DiFBench} benchmark creation framework, to enable systematic evaluation of collection-level statistical reasoning across domains (resumes, news summaries), document scales (10-40), and distinctiveness thresholds (2.5\%-20\%).
    \item We conduct the first large-scale study revealing that even leading LLMs degrades significantly with scale, with 75.9\% of errors resulting from misidentifying frequent features as distinctive. This provides computational evidence of base rate neglect in LLM reasoning.
    \item We demonstrate that explicit verification prompting leads to a 65\% relative improvement in the F1 score, offering a practical mitigation while highlighting persistent limitations in multi-document comparative reasoning.
\end{enumerate}

\begin{figure*}[ht]
    \centering
    \includegraphics[width=.9\linewidth]{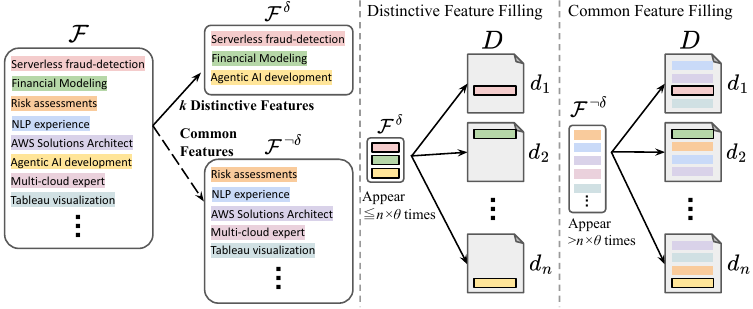}
    \caption{Overview of \textsc{DiFBench}. 
    To obtain distinctive features, $\mathcal{F}^\delta$, we first randomly select $k$ features from the feature set $\mathcal{F}$. The remaining features are treated as common features, $\mathcal{F}^{\neg\delta}$.
    Distinctive features $\mathcal{F}^\delta$ are distributed across documents while ensuring that each feature appears less than or equal to $\theta$\% of the documents. 
    Common features $\mathcal{F}^{\neg\delta}$ are then distributed across documents, ensuring that each feature appears in more than $\theta$\% of the $n$ documents.}
    \label{fig:overview}
\end{figure*}

\section{Related Work}
\label{sec:related}

\paragraph{Complex and Quantitative Reasoning in LLMs}
Recent benchmarks increasingly test multi-document reasoning, but their primary focus remains on aggregating query-relevant content or retrieving salient passages \cite{levy-etal-2024-task,bai-etal-2024-longbench,zhang-etal-2024-bench,hsieh2024ruler,bai2024longbench2,yen2025helmet,maekawa2025holistic}. In contrast, DFM shifts the focus to corpus-level statistical reasoning, requiring the identification of globally rare features. This requires reliable counting, base rate estimation, and population-level comparison. These are all areas where LLMs remain weak \cite{maekawa2025holistic}. Our findings reinforce this, showing that models often miscount feature frequencies and overestimate the distinctiveness of common traits. These limitations highlight statistical reasoning across documents as an underexplored and unresolved challenge.

\paragraph{Multi-document Summarization}

Multi-document summarization typically aims to synthesize common themes or provide a unified overview of content across documents \cite{li2012multisum,laban-etal-2024-summary,belem2024single}.
A few recent efforts \cite{huang-etal-2024-embrace,zhang-etal-2024-fair} have explored diversity-aware summarization, but they focus on maximizing coverage of perspectives rather than surfacing rare or distinctive features. DFM complements this line of work by targeting corpus-level rarity rather than within-document salience or inter-document consensus.

\paragraph{Comparative Summarization and Pairwise Analysis}
Prior work on comparative summarization has explored pairwise document contrast and entity differentiation \cite{iso22aclfindings,Gunel2023strum,gunel2024strumllm,yan-etal-2024-predicting}. These methods effectively highlight differences between two entities but do not scale to collections with many candidates. Crucially, they also lack a statistical frame for identifying what is rare relative to a population. DFM extends these efforts to multi-way comparisons, allowing models to reason over the distinctiveness of features in the context of an entire set—a key requirement in realistic decision-making scenarios such as hiring or product recommendation.

\section{Distinctive Feature Mining and Benchmark Creation Framework}
\label{sec:task}
We first introduce the task of Distinctive Feature Mining (DFM), present the design principles of \textsc{DiFBench}, a general benchmark creation framework designed to systematically evaluate models on this task. 
Then, we explain the details of the benchmark creation framework.
Finally, we describe how \textsc{DiFBench} is implemented to create benchmark datasets.

\subsection{Task Definition}
In this study, we simplify each document into a set of features. This can be realized by feature extraction methods \cite{clavie2023large} in common use cases such as resume screening and product comparison. 
Formally, a document set is denoted as $D = \{d_1, d_2, \ldots, d_n\}$, where each document $d_i$ consists of a set of up to $h$ features, $F_i = \{f^i_1, f^i_2, \ldots\}$. 
Let $\mathcal{F} = \bigcup_{i=1}^n F_i$ denotes the set of all features across $D$.
The task of Distinctive Feature Mining (DFM) is to identify, for each document $d_i$, a subset of features $F^\delta_i \subseteq F_i$ that are distinctive. A feature is considered \textit{distinctive} if it appears in at most $\theta$\%  of documents, where $\theta$ is a user-defined threshold.

\subsection{Design Principles}

We introduce \textsc{DiFBench}, a benchmark creation framework specifically designed to evaluate model performance on the DFM task. 
Figure~\ref{fig:overview} illustrates the overall process. Given a feature set $\mathcal{F}$, the framework partitions it into distinctive and common subsets, then distributes these features across documents $D$ in a controlled manner to enable systematic evaluation. \textsc{DiFBench} is guided by three core design principles:
\begin{enumerate}[label=(\arabic*), leftmargin=*]
    \item \textbf{Distinctive features in comparable candidates}: 
    Documents must be comparable; that is, all documents belong to the same domain and share the same structure. They differ in select features that make them distinctive.
    \item \textbf{Flexible number of candidates and distinctive features}: 
    Variable numbers of candidate documents and distinctive features must be supported to enable evaluation across scale and distinctiveness thresholds.
    \item \textbf{Systematic evaluation}: The framework must enable controlled experiments and facilitate precise measurement of model ability to detect globally rare features and reason over aggregate statistics.
\end{enumerate}

These principles enable a comprehensive testbed for studying corpus-level statistical reasoning. It allows researchers to probe models’ capacity to (1) extract features across documents, (2) count their frequencies, and (3) identify what is statistically distinctive in a given population.

\subsection{Benchmark Creation Framework}
\label{ssec:dataset_creation}

To realize these design principles in practice, \textsc{DiFBench} takes as input a set of features $\mathcal{F}$ and programmatically constructs a document set $D$ by distributing these features based on configurable parameters. These key parameters are:

\begin{enumerate}[label=(\arabic*), leftmargin=*]
  \item \textbf{Number of documents ($n$)}: Controls the scale of the dataset, allowing us to test how model performance varies with small to large document collections. Increasing $n$ raises the complexity of DFM as models must consider more candidates and interactions.
  \item \textbf{Number of distinctive features ($k$)}: Specifies how many features are truly distinctive across the document set. By varying $k$, we can simulate settings where distinctive traits are sparse or abundant, which affects the difficulty of mining such features.
  \item \textbf{Distinctiveness threshold ($\theta$)}: Defines the maximum proportion of documents a feature can appear in to be considered distinctive. This parameter enables us to influence feature rarity and overlap across documents.  
 
\end{enumerate}

Together, these provide fine-grained control over the complexity, sparsity, and overlap within the benchmark, enabling systematic and reproducible evaluation of statistical reasoning capabilities.

\paragraph{Document Set Construction}
The benchmark creation process begins by distributing distinctive features across a subset of documents, followed by populating the remaining feature slots with common features.
To this end, we first randomly select $k$ features from the set of features, $\mathcal{F}$, to serve as distinctive features, $\mathcal{F}^\delta$. Each distinctive feature is assigned a target document frequency, randomly sampled from the range $[1, n \times \theta]$, ensuring these features appear in only a small portion of the $n$ documents.
These distinctive features are then distributed across the documents to match their assigned frequencies.
The remaining features $F^{\neg\delta} = \mathcal{F} \setminus \mathcal{F}^\delta$ are treated as common features. Each is assigned a higher document frequency, sampled from the range $[n \times \theta + 1, n]$, and distributed across documents in the same way.
During assignment, we enforce a constraint that each document can contain at most $h$ features.
If a feature cannot be assigned without violating this rule, its assignment is skipped. This ensures that distinctive features remain relatively rare within the document set, while common features are broadly shared, thus preserving the intended distinction between the two categories.

\subsection{Benchmark Implementation}
\label{ssec:dataset_implementation}
While \textsc{DiFBench} is designed to accept any set of features, our implementation focuses on synthesizing features grounded in real-world source documents. Rather than relying on exact feature extraction, we opt for feature synthesis to support systematic and controlled evaluations. This approach ensures the generated features remain realistic and representative of the original documents while allowing us to precisely control task complexity.

\paragraph{Data Domains}
We use two different domains: resumes and news summaries—both well-suited for comparative analysis. 
For resumes, we source job posts from  \texttt{mycareerfuture.sg},\footnote{We downloaded the dataset from \url{https://github.com/WING-NUS/JD2Skills-BERT-XMLC}} selecting the 10 longest descriptions from each of five major job categories based on US Department of Labor statistics,\footnote{Labor Force Statistics from the Current Population Survey: \url{https://www.bls.gov/cps/cpsaat11.htm}}. These include {computer \& math, life physical \& social science, legal, architecture \& engineering, and healthcare occupations}. 
For news summaries, we utilized news articles from \cite{huang-etal-2024-embrace}, which cover five distinct topics. Each topic has 10 news articles. In summary, we have 100 source documents in total, with 50 resumes and 50 news summaries.

\begin{table*}[ht]
  \centering
  \resizebox{\linewidth}{!}{
  \begin{tabular}{lll}
      \toprule
      \textbf{Domain} & \textbf{Section} & \textbf{Synthesized Feature} \\
      \midrule
      Resume & Experience & Architected multi-cloud application frameworks aligning with banking industry compliance. \\
      & Technical Skills & Proficient in .NET Framework and .NET Core architectures \\
      & Soft Skills & Facilitated transparent communication across technical and non-technical audiences \\
      & Projects & Built serverless fraud-detection prototype leveraging AWS Lambda streams \\
      & Certifications & Achieved AWS Solutions Architect – Professional certification \\
      & Awards and Recognition & Earned Global Cloud Excellence award for innovative platform design \\
      \midrule
      News & fuel requirements & Inadequate ethanol content could trigger knock sensors and limp-home modes, ruining track sessions. \\
      Summary& vehicle performance & Carbon-fiber rim option trims 32 pounds of unsprung mass, quickening initial acceleration. \\
      & historical context & Factory 1,000-hp rating revives 1960s “horsepower wars” in a final escalation. \\
      & optional features & \$10,000 sunroof pricing intentionally discourages extra roof weight. \\
      & NHRA regulations & Street-legal Demons may drive NHRA to revisit Advanced ET class definitions. \\ 
      & production details & Compressed 2023 build window heightens risk of missed quotas before Brampton plant closure. \\ 
      & branding and marketing & Devilish \$96,666 base price turns MSRP into instant viral talking point. \\
      \bottomrule
  \end{tabular}
  }
  \caption{Examples of synthesized features in the resume and news summary dataset.}
  \label{tab:feature_examples}
\end{table*}

\paragraph{Feature Set Generation}
For each source document, we synthesize a set of features that are both relevant to its content and representative of its context. To guide feature creation, we use domain-specific structural templates. For resumes, these include categories such as Experience, Technical Skills, Soft Skills, Projects, Certifications, and Awards. For news summaries, we adopt 7–9 subtopics from the original dataset (e.g., \texttt{fuel requirements} under the motor trend topic) as section headers.

For each section, we prompt an LLM to generate a pool of 20 thematically relevant candidate features, using the seed document and section title as context. To encourage diversity across sections, we also supply the model with previously generated features from other sections of the same document. This helps ensure that each section’s features are both semantically relevant and distinct. We employ o3 \cite{openai2025o3o4mini} for a feature generation. Table \ref{tab:feature_examples} illustrates several examples of synthesized features. 

\section{Experimental Setup}
\label{sec:setup}

This section outlines our methodology for evaluating the statistical reasoning capabilities of LLMs. We first describe the parameters used for generating the synthetic document collections using \textsc{DiFBench}. We then introduce the suite of LLMs evaluated in our experiments, followed by a description of our inference and evaluation setup.

\paragraph{Document Set Construction Parameters} 
We set the number of documents $n$ to 10, 20, and 40 and test with $\lfloor n/2\rfloor$ distinctive features, to examine how LLMs handle varying levels of complexity in identifying distinctive features. 
We set the distinctive threshold $\theta$ to 2.5\%, 5\%, 10\%, and 20\% of the total documents (i.e., 1, 2, 4, and 8 documents respectively when $n=40$).
We set the maximum number of features per document $h$ to $4 \times S$, where $S$ denotes the number of sections of the document.

\paragraph{Models}
We evaluate 10 LLMs with reasoning-optimized and general-purpose capabilities. Reasoning models include both closed and open models: o3, o4-mini, Gemini-2.5-Flash, Qwen3-235B22A (Qwen3 for short). General models include GPT-4o, GPT-4o-mini, Gemini-2.5-Flash w/o think, Qwen3 w/o think, Llama-4-Maverick, and Llama-4-Scout. The model details are summarized in Appendix \ref{app:sec:models}. 
We set temperature and top-p parameters to $0.0$ and $1.0$, respectively, for all our experiments.

\paragraph{Inference Setup}
At inference time, each model is presented with a collection of documents generated by \textsc{DiFBench} and tasked with identifying the set of distinctive features within that collection. The model receives a single instruction prompt that asks it to return the features that appear rarely (distinctive features) for each document. Because \textsc{DiFBench} controls the construction of documents and explicitly selects which features are to be distinctive, we have access to ground-truth annotations $\mathcal{F}^\delta$ for each synthetic benchmark instance. This setup allows for objective evaluation of model predictions against a known gold standard.
We use the same prompt for all models, see Appendix \ref{app:sec:prompts}.

\paragraph{Evaluation Metrics}
Model predictions are evaluated using exact string match against the ground-truth set $\mathcal{F}^\delta$ provided by \textsc{DiFBench}. Our primary evaluation metric is the F1 score, with precision and recall reported in detailed analyses. 

\begin{figure*}[t]
    \centering
    \begin{subfigure}[b]{.9\linewidth}
        \centering
        \includegraphics[width=1.\linewidth]{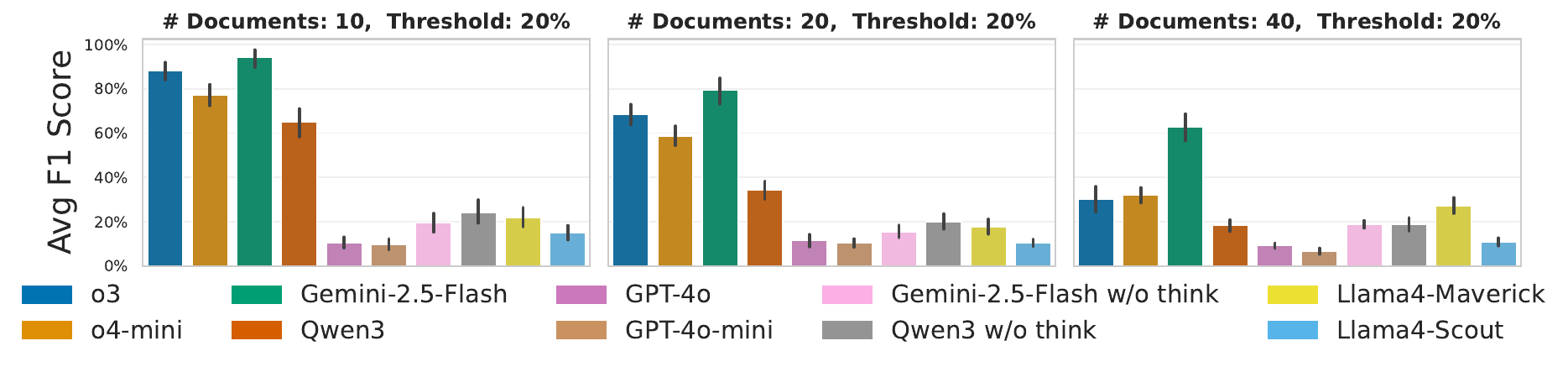}
        \caption{Resumes}
    \end{subfigure}
    \centering
    \begin{subfigure}[b]{.9\linewidth}
        \centering
        \includegraphics[width=1.\linewidth]{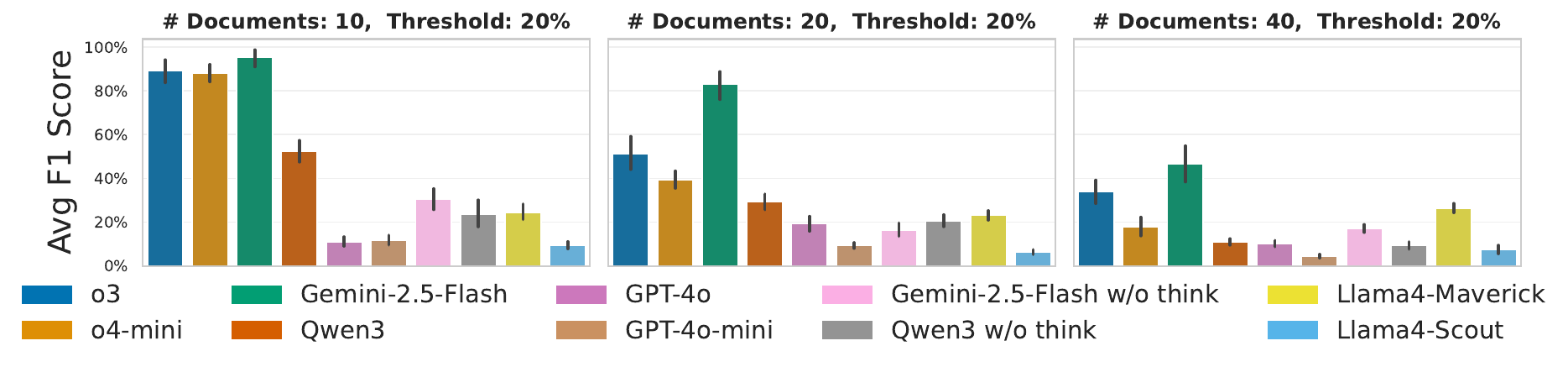}
        \caption{News summaries}
    \end{subfigure}
    \caption{F1 scores with various document sizes. The error bars indicate the standard deviation across samples. }
    \label{fig:various_document_sizes}
\end{figure*}
\begin{table}[h]
\resizebox{.95\linewidth}{!}{
    \begin{tabular}{l|cc}
    \toprule
    \textbf{Models} & \textbf{Resumes} & \textbf{News Summaries} \\
    \midrule
    o3 & 68.95 & 69.81 \\
    o4-mini & 61.92 & 58.45 \\
    Gemini-2.5-Flash & 84.78 & 77.76 \\
    Qwen3 & 46.41 & 36.32 \\
    \midrule
    GPT-4o & 12.55 & 17.12 \\
    GPT-4o-mini & 8.45 & 7.77 \\
    Gemini-2.5-Flash w/o think& 20.38 & 22.29 \\
    Qwen3 w/o think& 24.08 & 18.72 \\
    Llama4-Maverick & 25.89 & 25.34 \\
    Llama4-Scout & 11.87 & 7.21 \\
    \bottomrule
    \end{tabular}
    }
    \caption{Average F1 scores of the reasoning-optimized and general-purpose models on the DFM task across three document sizes, i.e., 10, 20, and 40, and two distinctive features, i.e., 10\% amd 20\%. The models with \textit{w/o} suffix are general models that do not use reasoning capabilities.}
    \label{tab:overall_results}
\end{table}

\section{Results and Analysis}
\label{sec:experiments}

\subsection{Main Results}
\label{ssec:results_overall}

\paragraph{Reasoning models consistently outperform their general counterparts. }
Table \ref{tab:overall_results} shows average F1 scores on the DFM task across varying document counts (10, 20, 40) and distinctiveness thresholds ($10\%$ and $20\%$). Overall, reasoning models consistently outperform general-purpose models across all settings. Surprisingly, no general model achieves F1 higher than $0.3$\%,  
indicating their limitation in identifying distinctive features effectively. This is particularly evident when comparing Gemini-2.5-Flash and Qwen3 with their non-reasoning (`w/o think') variants, where reasoning-optimized versions consistently perform better. This trend holds across both domains, resumes and news summaries.

\paragraph{Even current reasoning models are poor statistical reasoners when the collection size increases.}
To investigate the impact of number of documents on DFM  performance, we break down the results by document size in Figure \ref{fig:various_document_sizes}, focusing on the $20\%$ distinctive threshold. Results for $10\%$ threshold are included in the Appendix \ref{app:sec:results_distinctive_threshold_10}, where we observe similar trends.

Reasoning models consistently outperform general models across all document sizes, with F1 scores generally degrading as the number of documents increases. Their advantage is most evident with smaller sets (10 documents), where models like o3 and Gemini-2.5-Flash achieve F1 scores above $85\%$. However, performance drops sharply as the number of documents increases, with F1 scores dropping below $60\%$ for 40 documents in most cases. This suggests that while reasoning capabilities significantly benefit DFM, current models still struggle with the multi-document comparison at larger scales.

\begin{figure*}[t]
    \centering
    \begin{subfigure}[b]{0.48\linewidth}
        \centering
        \includegraphics[width=\linewidth]{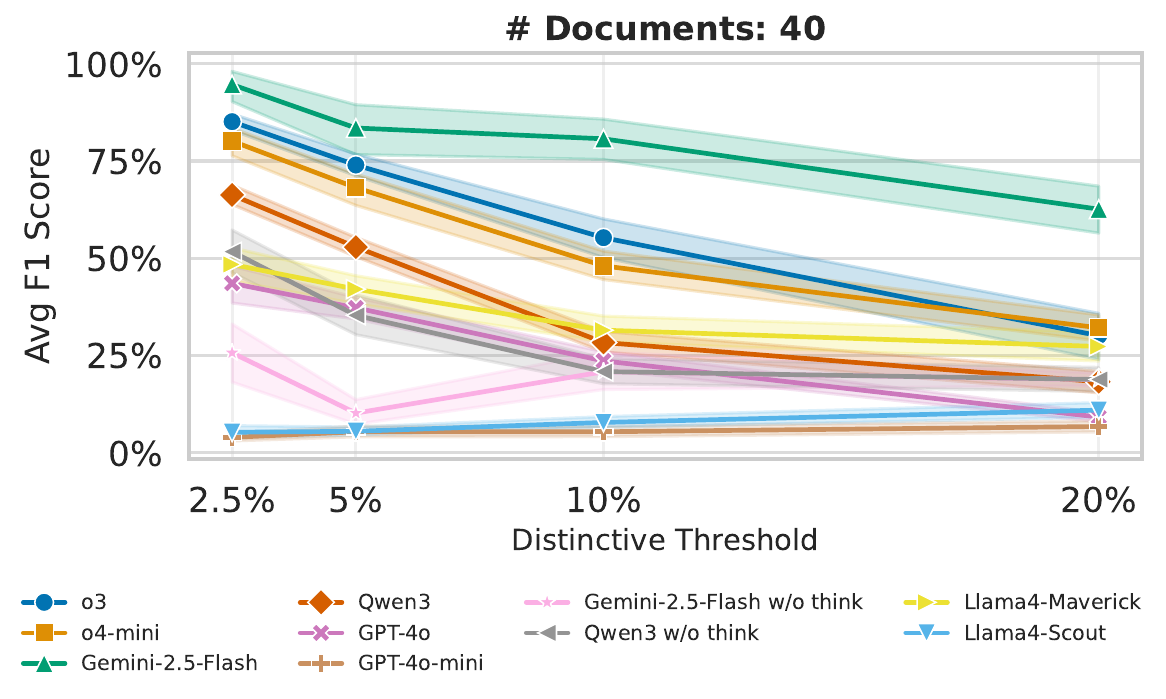}
        \caption{Resumes}
        \label{fig:f1_resumes}
    \end{subfigure}
    \hfill 
    \begin{subfigure}[b]{0.48\linewidth}
        \centering
        \includegraphics[width=\linewidth]{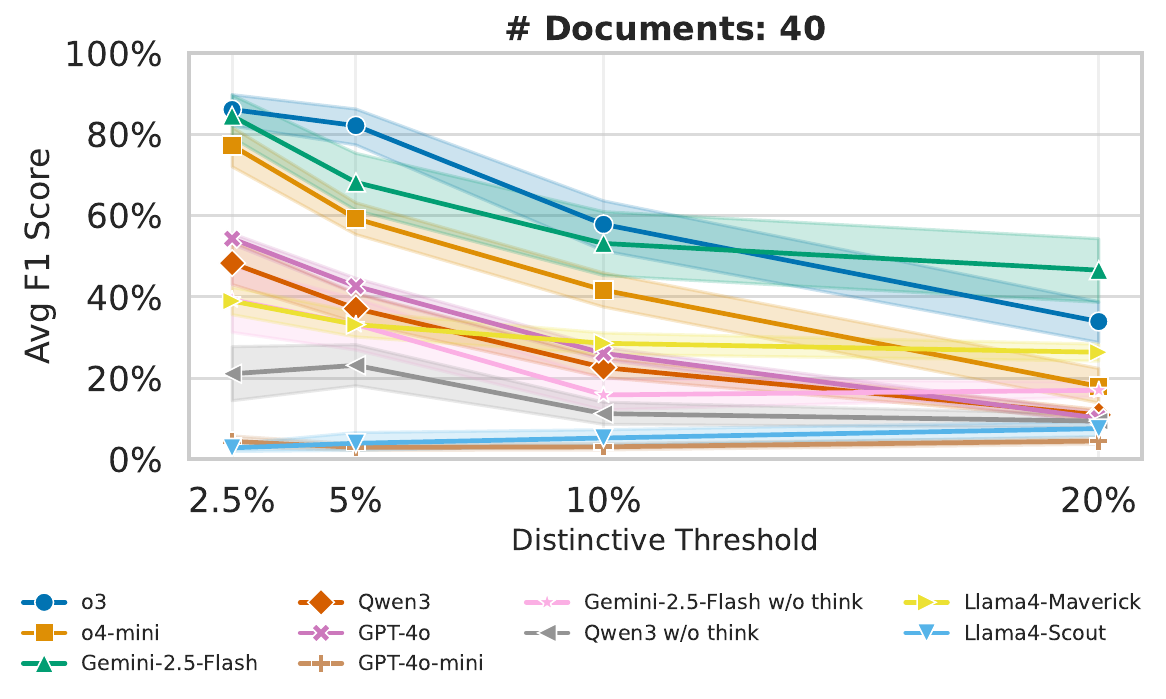}
        \caption{News summaries}
        \label{fig:f1_news}
    \end{subfigure}
    \caption{F1 scores with $40$ documents and various distinctive thresholds.}
    \label{fig:various_thresholds}
\end{figure*}

\paragraph{Statistical reasoning becomes more challenging as the distinctive threshold increases.}
We further analyze F1 scores across varying distinctive thresholds, keeping the number of documents fixed at 40 (see Figure \ref{fig:various_thresholds}). We observe that F1 scores generally decline as the threshold increases, suggesting it becomes harder for models to isolate features that distinguish fewer documents from a larger set. At higher thresholds (e.g., 20\%), all models perform poorly, with narrow gap between reasoning and general models. This implies that finer-grained DFM still remains a key challenge even for advanced reasoning models.

\begin{figure}[t]
    \centering
    \includegraphics[width=.9\linewidth]{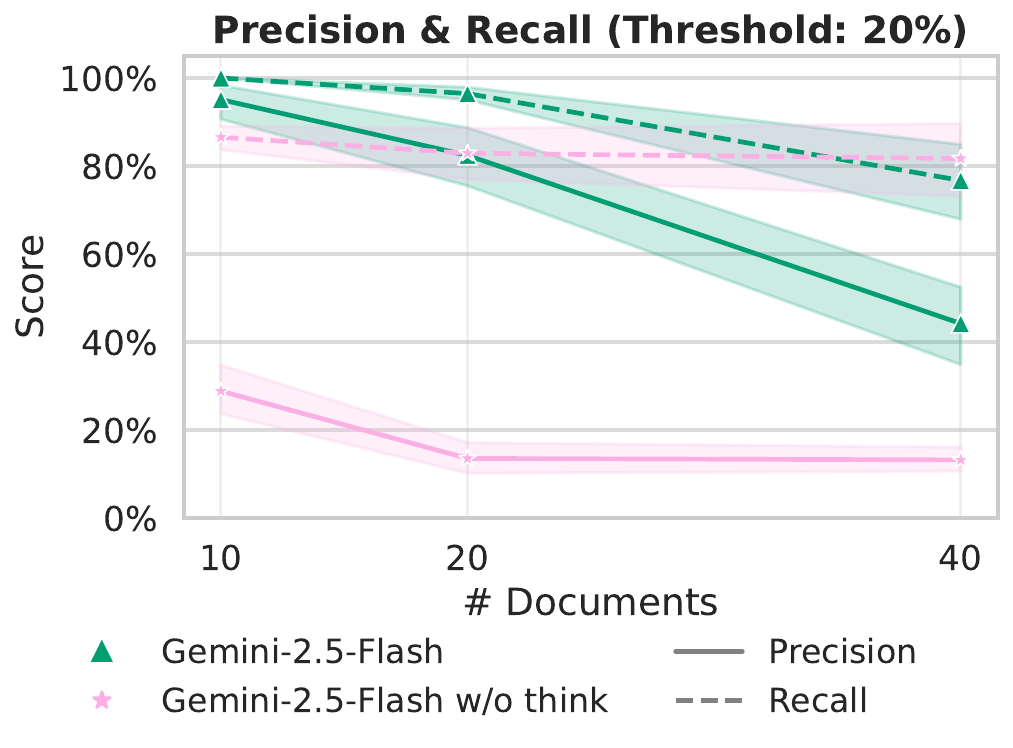}
    \caption{The precision and recall}
    \label{fig:precision_recall}
\end{figure}

\subsection{Analysis of DFM Performance}

We conduct a deeper analysis of LLM performance on the DFM task by examining precision, recall and token usage. Due to space constraints,, we present detailed results on the news summary dataset here and include additional results on the resume dataset in Appendix \ref{app:sec:precision_recall_resume}, which follow similar trends.

\paragraph{While general models tend to over-predict, reasoning models are more selective in identifying truly distinctive features.}
Figure \ref{fig:precision_recall} shows the average precision and recall of Gemini-2.5-Flash and its w/o think variant across various document sizes.
Gemini-2.5-Flash generally achieves higher precision than its w/o think, suggesting that reasoning models are more effective at identifying truly distinctive features. 
Interestingly, general models achieve relatively higher recall but poor precision (<30\%), suggesting they tend to over-predict and include many irrelevant features. This reflects a lack of selectivity in general models when attempting feature mining under increasing complexity.
Finally, we observe that both precision and recall drop as the document size increases even for the best-performing reasoning model, Gemini-2.5-Flash. This indicates that as the number of documents increases, models struggle to accurately count feature occurrences and identify those that are truly distinctive.

\begin{figure}[t]
    \centering
        \includegraphics[width=1.\linewidth]{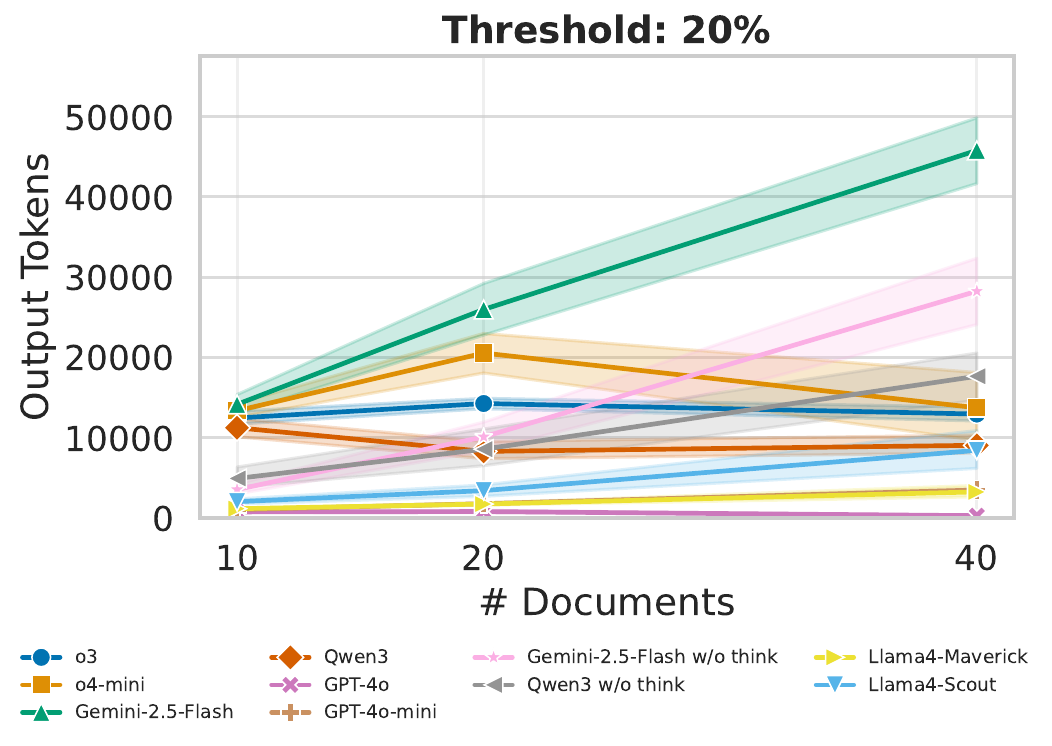}
    \caption{The average number of output tokens.}
    \label{fig:token_usage}
\end{figure}

\paragraph{Better statistical reasoning requires more output tokens.}
Figure \ref{fig:token_usage} shows the average number of output tokens under varying document sizes. 
As shown, reasoning models tend to generate more tokens in total as the number of documents increases. 
This indicates that models require more reasoning to identify distinctive features when the complexity of the task increases. 
Notably, when considered alongside the results in Figure \ref{fig:various_document_sizes}, Gemini-2.5-Flash achieves a high F1 score by significantly increasing its token usage compared to other models. 
We also observe that Gemini-2.5-Flash w/o think generates a larger number of tokens than most of other models, despite its low precision score (see Figure \ref{fig:precision_recall}).
This suggests that the model struggles with the statistical reasoning even if it generates a large amount of tokens to identify rare information. 
The results on the resume dataset is included in Appendix \ref{app:sec:token_usage_resume}, which shows similar trends.

\subsection{Error Analysis}
\label{sssec:error_analysis}
To better understand model limitations, we analyze the errors made by the best-performing model, Gemini-2.5-Flash, in the most challenging setting (40 documents and a distinctive threshold of 20\%). We categorize the errors into three main types: (1) \textbf{Non-distinctive.} Features that are mentioned in the document but are not distinctive. (2) \textbf{Contamination.} Features that are not mentioned in the document itself but occur in other documents. (3) \textbf{Typo/Abbreviation.} Features that are not mentioned in any documents, often due to typos or malformed abbreviations.

\paragraph{The best performing model still struggles to estimate frequencies of features.}
Table \ref{tab:category_distribution} shows the distribution of these errors. 
The majority of errors are non-distinctive features, which indicates that the model tends to identify features that appear in the document but are not truly distinctive.
This result suggests that models struggle to correctly estimate the frequencies of features if they handle many features.
We also observe that the model makes contamination errors, which indicates that the model tends to identify features that are not mentioned in the document but are present in other documents.
Since the recent model follows the instruction well, we observe that the model rarely makes typo/abbreviation errors.

\begin{table}[t]
\centering
  \resizebox{0.75\linewidth}{!}{
\begin{tabular}{lr}
\toprule
Category & Percentage (\%) \\
\midrule
Non-distinctive & 75.90 \\
Contamination & 1.89 \\
Typo/Abbreviation & 0.01 \\
\midrule
Correct & 22.20 \\
\bottomrule
\end{tabular}
}

\caption{The distribution of error categories.}
\label{tab:category_distribution}
\end{table}

\subsection{Mitigation Strategy}
Motivated by the observations that models achieve high precision but low recall in Figure \ref{fig:precision_recall} and the majority of errors comes from non-distinctive features in Table \ref{tab:category_distribution}, we propose a simple post-processing strategy to mitigate the errors.
The mitigation strategy is as follows: 1) for each model-generated feature, we have the model judge whether the feature is distinctive or not by comparing the feature with all documents, and 2) if the feature is distinctive, we keep it; otherwise, we discard it.
This strategy is based on the assumption that if the model processes features one by one, it can better identify whether the feature is distinctive or not than processing all features at once. 

\begin{figure}[t]
    \centering
    \includegraphics[width=.9\linewidth]{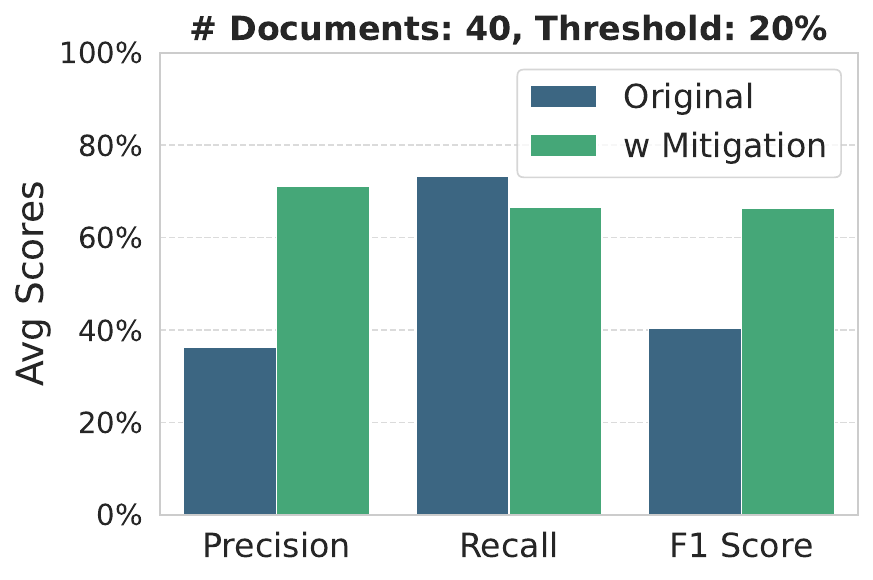}
    \caption{Effect of our mitigation strategy. }
    \label{fig:mitigation}
\end{figure}

\paragraph{The mitigation strategy dramatically improves the precision of the DFM task but it is still far from perfect.}
Figure \ref{fig:mitigation} shows the F1 scores of the DFM task with and without the mitigation strategy. We use Gemini-2.5-Flash as a judge model and other settings are the same as Table \ref{tab:category_distribution}.
We observe that the mitigation strategy successfully improves the precision with a slight decrease in recall, resulting in a higher F1 score which is a $65\%$ relative improvement over the original score. 
However, the mitigation strategy requires high cost and adds latency, as it needs to compare each model-generated feature with all documents to check whether the feature is distinctive or not. Also, while the mitigation strategy improves the precision, it is still around $70\%$, which indicates that the model still struggles to identify distinctive features correctly even if it processes features one by one.
We leave the exploration of more efficient and reliable mitigation strategies for future work.

\section{Conclusion}
\label{sec:conclusion}
We introduced \textsc{DiFBench}, a configurable benchmark creation framework for the distinctive feature mining (DFM) task, designed to systematically evaluate the statistical reasoning capabilities of LLMs.
Through extensive experiments with ten models—four reasoning and six general-purpose—we found that reasoning models consistently outperform general ones but degrade sharply as the number of documents increases. Our analysis revealed that even strong models struggle with precision and often misidentify common features as distinctive.
While our mitigation strategy substantially improved precision, it also emphasized the need for more robust methods to support reliable statistical reasoning in complex multi-document settings.

\section*{Limitations}
\label{sec:limitations}

While this work systematically evaluates how LLMs identify distinctive single feature, a logical next step is to explore combinational distinctiveness. Examining unique combinations of features would more accurately reflect the complexity of real-world scenarios.

Since the purpose of our work is to assess the capabilities of LLMs in statistical reasoning, we simply evaluate every feature in the document set without considering their weight or importance.
The evaluation framework does not account for the weight of features, which can be a significant factor in determining their importance. In real-world scenarios, some features may carry more weight than others, influencing their relevance and distinctiveness. For instance, in a resume analysis context, certain skills or experiences may be more critical than others. Future work could explore methods to incorporate feature weighting into the evaluation process, allowing for a more nuanced assessment of the distinctive feature mining.

Additionally, our current evaluation relies on an exact string match, which simplifies the task. A natural next step for future work is to increase the complexity by incorporating paraphrased features. This would require models to identify semantically equivalent but textually different features, making the benchmark more challenging and aligned with real-world complexities.

\section*{Ethical considerations}
\label{sec:ethical_considerations}
We acknowledge the ethical implications of Distinctive Feature Mining (DFM), particularly in high-stakes domains like hiring. While our work is a technical exploration of statistical reasoning, its application requires careful foresight. Key concerns include:

\begin{itemize}
    \item \textbf{Bias Amplification and Proxy Discrimination:} DFM identifies statistically rare features without semantic understanding. This risks flagging features that are proxies for protected attributes (e.g., race, gender, age), potentially amplifying societal biases in downstream applications.

    \item \textbf{Novelty vs. Competency:} The task's focus on rarity may lead to prioritizing novel features over core competencies. This could undervalue well-qualified candidates with standard skill sets in favor of those with unique but less relevant attributes.

    \item \textbf{Reductionism and Dehumanization:} A feature-centric view is inherently reductionist, simplifying complex entities like candidates into a list of keywords. This risks a dehumanizing evaluation process that overlooks holistic qualities like critical thinking or creativity.
\end{itemize}

Future work must address these risks, for instance by developing fairness-aware frameworks that can distinguish between meaningful and potentially discriminatory features. We release our benchmark to encourage community research into both the capabilities and societal risks of this technology.

While we used AI assitants such as ChatGPT and Copilot to assist in coding and revising this paper, we carefully reviewed and edited all content to ensure it meets our standards and aligns with our research goals.

\bibliography{anthology,custom}

\appendix

\begin{figure*}[t]
    \centering
    \begin{subfigure}[b]{.9\linewidth}
        \centering
        \includegraphics[width=1.\linewidth]{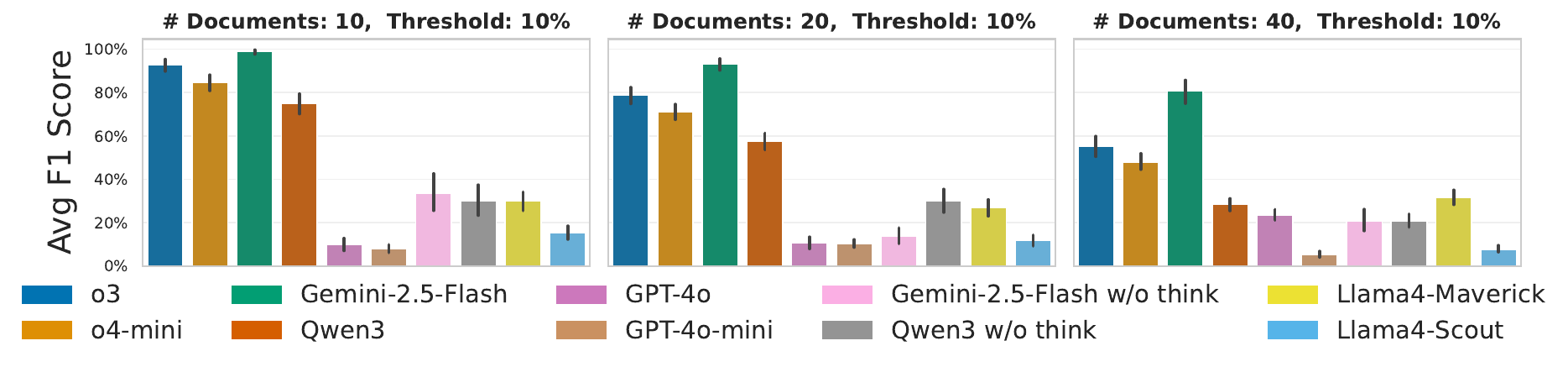}
        \caption{Resumes}
    \end{subfigure}
    \centering
    \begin{subfigure}[b]{.9\linewidth}
        \centering
        \includegraphics[width=1.\linewidth]{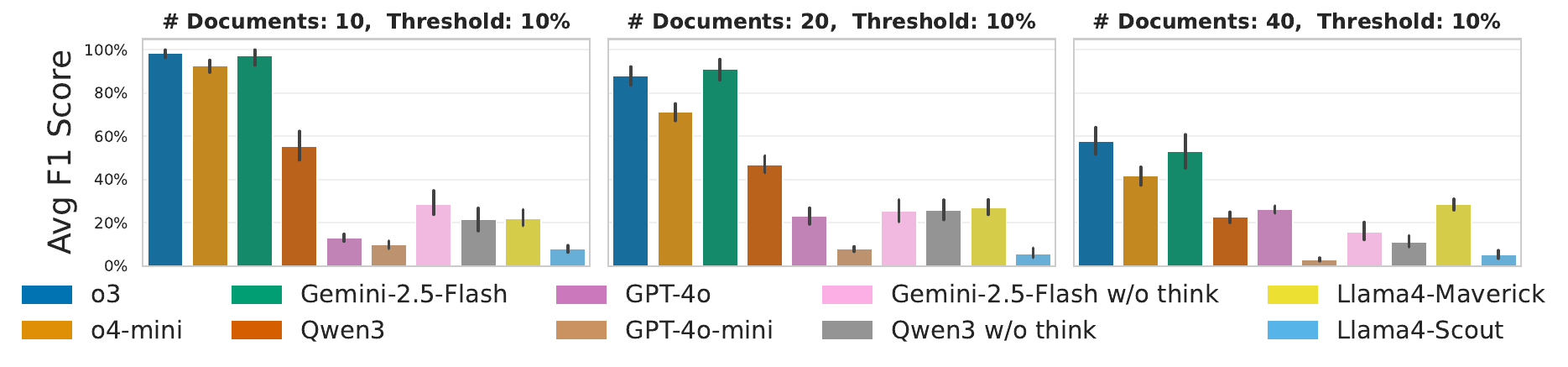}
        \caption{News summaries}
    \end{subfigure}
    \caption{F1 scores with various document sizes when $\theta=0.1$.}
    \label{fig:overall_results_10}
\end{figure*}
\begin{figure}[t]
    \centering
    \includegraphics[width=.9\linewidth]{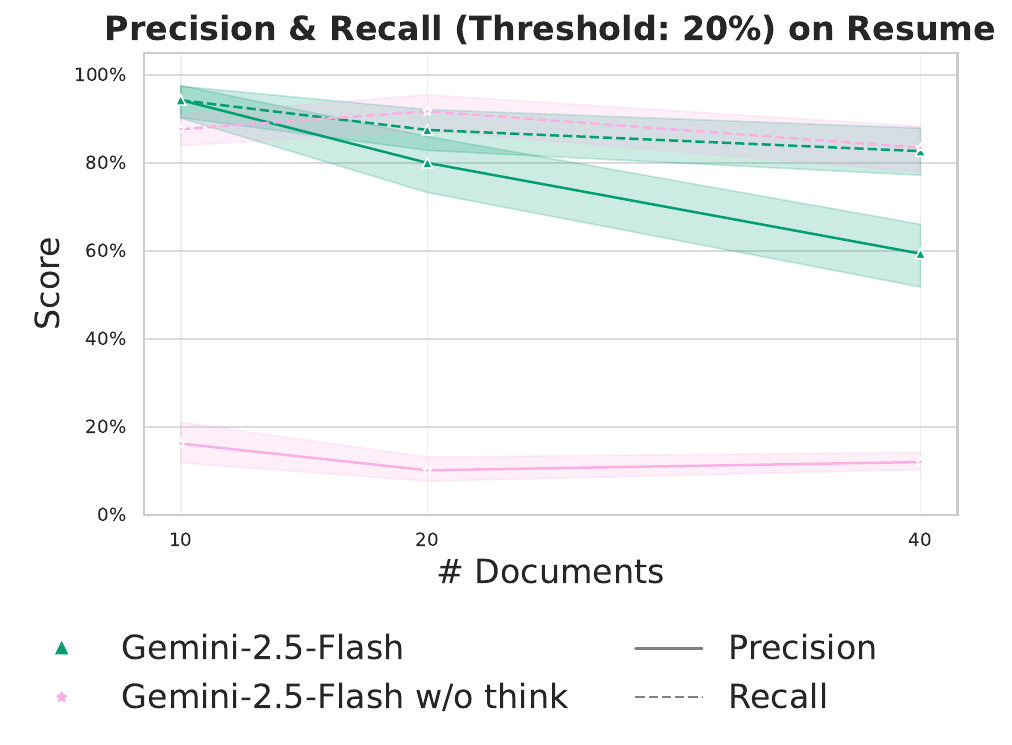}
    \caption{The precision and recall with various document sizes (Resume dataset).}
    \label{app:fig:precision_recall}
\end{figure}
\begin{figure}[t]
    \centering
    \includegraphics[width=.9\linewidth]{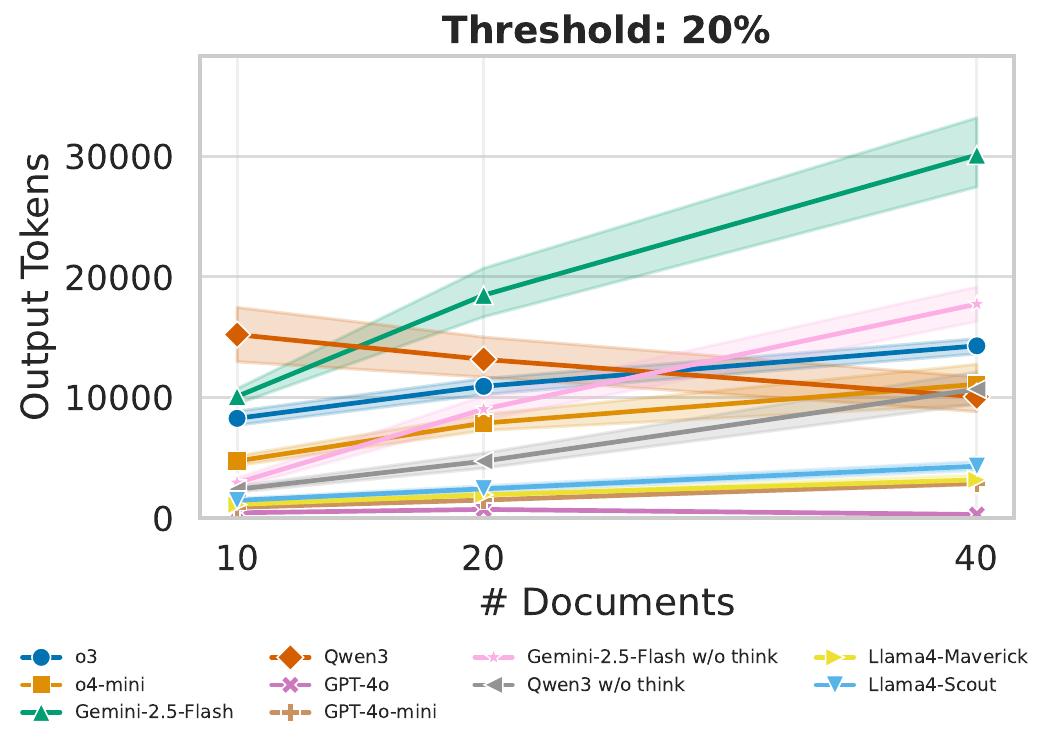}
    \caption{The average number of output tokens with various document sizes (Resume dataset).}
    \label{app:fig:token_usage}
\end{figure}

\section{Additional Results}
\label{app:sec:additional_results}
In this section, we present additional results on the resume dataset to complement the main results presented in the paper.

\subsection{Results with Distinctive Threshold 10\%}
\label{app:sec:results_distinctive_threshold_10}

Figure \ref{fig:overall_results_10} shows the F1 scores with various document sizes when the distinctive threshold is set to $10\%$. 
Overall, the results follow similar trends to those with $20\%$ threshold (Figure \ref{fig:various_document_sizes}), 
i.e., reasoning models outperform general-purpose models especially when the document size is small.
Aslo, we observe that the F1 scores are generally higher than those with 20\% threshold (Figure \ref{fig:various_document_sizes}), 
indicating that it is easier for models to identify distinctive features when the threshold is lower. 
This corresponds to our findings in Figure \ref{fig:various_thresholds}, where the F1 scores generally decrease as the distinctive threshold increases.

\subsection{Precision and Recall on Resume Dataset}
\label{app:sec:precision_recall_resume}

Figure \ref{app:fig:precision_recall} shows the precision and recall of Gemini-2.5-Flash and its w/o think variant on the DFM task with various document sizes when the distinctive threshold is set to $20\%$.
We observe that the trends are similar to those in Figure \ref{fig:precision_recall} for the news summary dataset.
While Gemini-2.5-Flash achieves higher precision than its w/o think variant, the precision declines as the document size increases.

\subsection{Token Usage on Resume Dataset}
\label{app:sec:token_usage_resume}

Figure \ref{app:fig:token_usage} shows the average number of output tokens with various document sizes when the distinctive threshold is set to $20\%$.
We observe a trend that the number of output tokens increases as the document size increases, similar to the trend observed in Figure \ref{fig:token_usage} for the news summary dataset.

\begin{table*}[h]
  \centering
  \resizebox{\linewidth}{!}{
  \begin{tabular}{lrrll}
      \toprule
      \textbf{Model} & \textbf{Size} & \textbf{Context} & \textbf{HuggingFace / API} & \textbf{License}\\
      \midrule
      o3 \citep{openai2025o3o4mini} & - & 200k & \texttt{o3-2025-04-16} & OpenAI Service Terms\footnotemark[1] \\
      o4-mini \citep{openai2025o3o4mini} & - & 200k & \texttt{o4-mini-2025-04-16} & OpenAI Service Terms\\
      Gemini-2.5-Flash \citep{comanici2025gemini} & --- & 1M & \texttt{gemini-2.5-flash-preview-04-17} & Gemini API Additional Terms of Service\footnotemark[2]\\
      GPT-4o \citep{openai2024gpt4o}& --- & 128k & \texttt{gpt-4o-2024-08-06} & OpenAI Service Terms \\
      GPT-4o-mini \citep{openai2024gpt4o}& --- & 128k & \texttt{gpt-4o-mini-2024-07-18} & OpenAI Service Terms\\
      Llama-4-Maverick \citep{meta2025llama4} & 400B & 1M & \texttt{meta-llama/Llama-4-Maverick-17B-128E-Instruct} & Llama 4 Community License Agreement\footnotemark[3] \\ 
      Llama-4-Scout \citep{meta2025llama4} & 109B & 10M & \texttt{meta-llama/Llama-4-Scout-17B-16E-Instruct} & Llama 4 Community License Agreement \\
      Qwen-3 \citep{qwen3technicalreport} & 235B & 128k & \texttt{Qwen/Qwen3-235B-A22B} & Apache license 2.0 \\
      \bottomrule
  \end{tabular}
  }
  \caption{Models used in experiments. Model sizes are not publicly disclosed (-).}
  \label{tab:models}
\end{table*}

\section{Model Details}
\label{app:sec:models}

Table \ref{tab:models} shows the model details used in our experiments.

\footnotetext[1]{\url{https://openai.com/policies/services-agreement/} [Accessed: July 26, 2025]}
\footnotetext[2]{\url{https://ai.google.dev/gemini-api/terms} [Accessed: July 26, 2025] }
\footnotetext[3]{\url{https://www.llama.com/llama4/license/} [Accessed: July 26, 2025]}

\section{Prompts}
\label{app:sec:prompts}

Figure \ref{fg:app:task_prompt} shows a prompt used for the DFM task in this paper.

\begin{figure*}[t]
  \begin{tcolorbox}
  \footnotesize
\textbf{\# Role} \\
You are an expert AI assistant specializing in comparative resume analysis.

\vspace{1em}
\textbf{\# Objective} \\
You will be given \{num\_documents\} resumes. For each candidate, identify their "distinctive features" (such as skills, tools, or certifications) that are held by \{distinctive\_threshold\} or fewer of the total candidates.

\vspace{1em}
\textbf{\# Instructions}
\begin{itemize}[nosep]
    \item Identify features from each resume (e.g., programming languages, software tools, professional certifications, unique projects).
    \item Provide your thinking and reasoning process before listing the features.
    \item Count the occurrences of each feature across all resumes to determine which ones meet the "distinctive" criteria (appearing \{distinctive\_threshold\} or fewer times).
    \item For each candidate, create a list of the distinctive features they possess.
    \item When listing features, use the exact wording as it appears in the resume. Do not summarize or rephrase.
    \item If a candidate has no qualifying distinctive features, return an empty list `[]`.
    \item Your output must be in valid JSON format.
\end{itemize}

\vspace{1em}
\textbf{\# Input Resumes:}
\begin{verbatim}
{resumes}
\end{verbatim}

\textbf{Output Format} (\{num\_documents\} candidates)
\begin{verbatim}
{{
  "outputs": [
    {{
      "candidate_id": 1,
      "reasoning": "Your reasoning and analysis for candidate 1",
      "output": [
        distinctive_feature_1,
        distinctive_feature_2,
        ...
        ]
    }},
    ...
    {{
      "candidate_id": {num_documents},
      "reasoning": "Your reasoning and analysis for candidate {num_documents}",
      "output": [
        distinctive_feature_1,
        distinctive_feature_2,
        ...
        ]
    }}
  ]
}}
\end{verbatim}
  \end{tcolorbox}
  \caption{DFM task prompt template for the resume dataset. Variables \{num\_documents\} and \{distinctive\_threshold\} are replaced with the number of documents and the distinctive threshold, respectively. The \{documents\} variable is replaced with the actual documents.}
  \label{fg:app:task_prompt}
\end{figure*}

\end{document}